\documentclass[runningheads]{llncs}
\usepackage{graphicx}
\usepackage{tikz}
\usepackage{enumerate}

\newcommand{\citeA}[1]{\cite{#1}}

\usetikzlibrary{positioning,chains}
\usepackage{subfig}
\usepackage{hyperref}

\begin{document}
\title{Investigating Efficient Learning and Compositionality in Generative LSTM Networks}
\titlerunning{Generative LSTM Networks and Compositionality}
%\author{Sarah Fabi\thanks{Corresponding author, {\email{sarah.fabi@uni-tuebingen.de}}} \orcidID{0000-0001-7149-7147}, Sebastian Otte \orcidID{0000-0002-0305-0463}, Jonas Gregor Wiese \orcidID{0000-0002-8925-2823}, Martin V. Butz \orcidID{0000-0002-8120-8537}}
\author{Sarah Fabi\thanks{Corresponding author, {\email{sarah.fabi@uni-tuebingen.de}}}, Sebastian Otte, Jonas Gregor Wiese, Martin V. Butz}

\authorrunning{S. Fabi et al.}
\institute{Neuro-Cognitive Modeling Group, Eberhard Karls University T\"ubingen, T\"ubingen, Germany\\
%\url{https://uni-tuebingen.de/fakultaeten/mathematisch-naturwissenschaftliche-fakultaet/fachbereiche/informatik/lehrstuehle/cognitive-modeling/}
}

\maketitle              % typeset the header of the contribution
\begin{abstract}	
When comparing human with artificial intelligence, one major difference is apparent:
Humans can generalize very broadly from sparse data sets because they are able to recombine and reintegrate data components 
in compositional manners. 
To investigate differences in efficient learning, Joshua B. Tenenbaum and colleagues developed the character challenge:   
First an algorithm is trained in generating handwritten characters.
In a next step, one version of a new type of character is presented.
An efficient learning algorithm is expected to be able to re-generate this new character, to identify similar versions of this character, to generate new variants of it, and to create completely new character types.
In the past, the character challenge was only met by complex algorithms that were provided with stochastic primitives.
Here, we tackle the challenge without providing primitives. 
We apply a minimal recurrent neural network (RNN) model with one feedforward layer and one LSTM layer and train it to generate sequential handwritten character trajectories from one-hot encoded inputs. 
To manage the re-generation of untrained characters when presented with only one example of them, we introduce a one-shot inference mechanism: 
the gradient signal is backpropagated to the feedforward layer weights only, leaving the LSTM layer untouched. 
We show that our model is able to meet the character challenge by recombining previously learned dynamic substructures, which are visible in the hidden LSTM states.
Making use of the compositional abilities of RNNs in this way might be an important step towards bridging the gap between human and artificial intelligence. 

\end{abstract}

\section{Introduction}
Despite numerous recent success stories with Deep Learning (DL) \cite{LeCun.2015} including game playing as well as image and speech recognition, over the last years various limitations in DL have been uncovered.
One important issue lies in the fact that DL algorithms lack mechanisms that lead to the development of hierarchical, generative structures in a natural way \cite{marcus2018deep}.
Current DL algorithms essentially learn correlations between features on a flat plane. 
When dealing with hierarchical problems, approximations are applied, which are often incorrect and do not generalize well.
As a result, DL algorithms are still easily fooled \cite{Nguyen:2015} and are not particularly or naturally noise-robust \cite{Geirhos:2018}.

Following the same line of reasoning, Lake and colleagues \citeA{lake2018still} stated that Recurrent Neural Networks (RNNs) do not develop compositional representations. 
That is, they are not able to parse an object or event into its components and flexibly recombine them in novel ways, making generalization hard, especially when training and test sets differ significantly.
A related phenomenon was also analyzed by Otte and colleagues \citeA{otte2019}, showing that the compositional disentanglement of superimposed dynamics is only possible when additional inductive learning biases of modularization and error distribution are added to the standard backpropagation through time weight adaptation mechanism in RNNs.

In the light of these current DL deficiencies, one may state that there exist two different kinds of artificial intelligence (AI) systems: ones that are inspired by human cognition and ones that are not.
Current DL techniques are mostly of the second type. 
Accordingly, Marcus \citeA{marcus2018deep} and Lake et al. \citeA{lake2017building} propose that in order to overcome the flaws of current DL systems, researchers should apply human cognition as a model for AI systems. Marcus \citeA{marcus2018deep} advises against using findings of the human brain, since the insights of neuroscience are not yet advanced enough.
He even assumes that we will need advanced AI systems to understand the human brain in the future.
In his view, cognitive science and developmental psychology are more promising models than neuroscience, since they already provide helpful insights into human intelligence.

In line with this argument, Hassabis et al. \citeA{hassabis2017neuroscience}, Lake et al. \citeA{lake2018still}, and Marcus \citeA{marcus2018deep} identify several key areas of cognition, in which human intelligence still outperforms artificial intelligence. 
These include intuitive physics and folk psychology, imagination, reasoning, and planning, as well as learning efficiency.
Here we focus on imagination and efficient learning.    

Humans have mental models, which enable the anticipation of future outcomes based on experiences. 
As a result, actions can be chosen in explicitly goal-directed manners. 
Moreover, imaginations are compositional in nature, that is, we are able to recombine previous experiences in innovative but meaningful manners. 
Hassabis and colleagues \citeA{hassabis2017neuroscience} pose the challenge that DL algorithms should be enhanced such that generative models of the environment are developed that allow compositional, simulation-based planning -- ideally without handcrafting strong priors into the DL network architecture. 

Another main superiority of human compared to artificial intelligence is efficient learning \cite{hassabis2017neuroscience}: Humans but not DL algorithms generalize very broadly from a sparse amount of data \cite{hassabis2017neuroscience,lake2017building}. The resulting rich conceptual representations can then be applied to a wide range of tasks, like parsing an object into its components or generating new examples of a concept. 
They even allow the creative generation of novel concepts by putting components together in a new but somewhat meaningful manner \cite{hassabis2017neuroscience}. 
This ability to re-combine structures in a compositional manner is a very important ingredient of efficient human learning and productivity because a finite number of conceptual primitives can be recombined into a mere infinite number of instances. 
Lake et al. \citeA{lake2017building} argue that this enables the brain to think an infinite number of thoughts and understand an infinite number of sentences. 

To investigate efficiency differences between human and artificial learning, Lake et al. \citeA{lake2015human} developed the \emph{character challenge}. It investigates one-shot classification and generation of handwritten characters. Those are well-suited for investigation because they are two-dimensional, clearly separated from the background, and unoccluded \cite{lake2017building}. The character challenge consists of the following tasks, combining several fundamental AI challenges \cite{hofstadter1985metamagical}:

\begin{enumerate}[i.]
\item Free generation of characters (after a single example) 
\item Generation of new samples of a concept (after a single example)
\item Identifying novel instances of a concept (after a single example)
\item Generation of whole new concepts (after single examples of some concepts)
\end{enumerate}

Lake and colleagues \citeA{lake2015human} applied Bayesian program learning (BPL), representing concepts as stochastic programs to achieve results in machines that are comparably efficient to those generated by humans. Structure sharing across concepts was accomplished by re-using the components of stochastic motor primitives.
The motor primitives were handcrafted and provided as priors to the BPL architecture. 

Following the demand of Hassabis et al. \citeA{hassabis2017neuroscience} to build simple models without handcrafted priors by the experimenter, in this paper, we aim at building a generative neural network model that faces the character challenge without providing handcrafted motor primitives. 
As a result, we are addressing efficient learning with respect to the character challenge, investigating to which extent imagination, planning, and compositional encodings and recombination abilities develop.
In particular, this paper introduces a generative RNN architecture that integrates the request for simulation-based planning, thereby managing the character challenge without prior information about stochastic primitives.
We show that our recurrent LSTM network, when trained on some characters, becomes able to recombine previously learned dynamical substructures when facing the task of generating untrained characters, of which only one variant is presented. 
With such compositional structures at hand, the network is not only able to re-generate those untrained characters, but it is also able to create new examples of a particular type of character and even totally new ones. 
Moreover, we show that the network is able to recognize similar variants of a (potentially just recently learned) character.

\section{Data and Model}
Handwritten characters of the Latin alphabet were collected from $10$ subjects, obtaining $440$ samples per character in total.
Each character trajectory was a sequence of a variable amount of time steps with two positional features that indicated the relative change in position in $x$ and $y$ direction, one feature representing the pressure with which the character was written and one representing the onset of a stroke.
The labels, which served as the input, consisted of one-hot encoded input vectors with length $26$ for every time step.  
The first $50 \%$ of the characters of the alphabet (a - m) were used to train the network, whereas only one variant of the remaining untrained characters (n - z) each was presented to the network during the different character challenge tasks. 

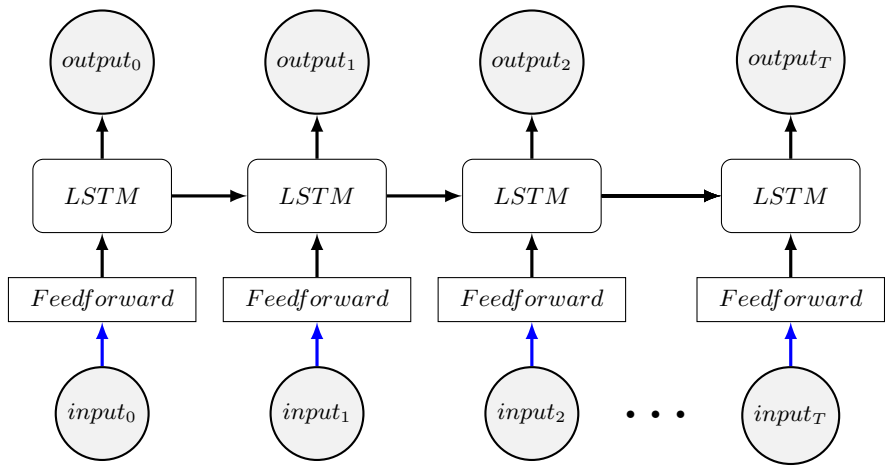
\begin{figure}[t!]
\centering
\begin{tikzpicture}[item/.style={circle, draw, thick,align=center}, 
itemff/.style={rectangle, draw,align=center, text width=7em, minimum height=1.8em, join},
itemc/.style={rectangle, rounded corners, draw ,on chain,join,align=center, text width=5em, minimum height=3em}]

 \begin{scope}[start chain=going right,nodes=itemc,every
 join/.style={-latex,very thick},local bounding box=chain]
 \path node (A0) {$LSTM$} node (A1) {$LSTM$} node (A2) {$LSTM$} node[xshift=1.8em] (AT)
 {$LSTM$};
 \end{scope}
 
 \begin{scope}[start chain=going right,nodes=itemff,every
 join/.style={-latex,very thick},local bounding box=chain] \path node[below = 1.8em of A0] (ff0) {$Feedforward$} node[below = 1.8em of A1] (ff1) {$Feedforward$} node[below = 1.8em of A2] (ff2) {$Feedforward$} node[below = 1.8em of AT] (ffT) {$Feedforward$};
 \end{scope} 
 
 \foreach \X in {0,1,2,T} 
 {\draw[very thick,-latex] (A\X.north) -- ++ (0,1.8em)
 node[above,item,fill=gray!10] (h\X) {$output_\X$};
 \draw[blue,very thick,latex-] (ff\X.south) -- ++ (0,-1.8em)
 node[black,below,item,fill=gray!10] (x\X) {$input_\X$};
 \path[very thick, -latex] (ff\X) edge (A\X);}
 
 \path (x2) -- (xT) node[midway,scale=2,font=\bfseries] {\dots};
\end{tikzpicture}
\caption{Model architecture: At every time step, the one-hot encoded input of length 26 is processed in a first fully-connected feed-forward layer with $100$ neurons, followed by one recurrent LSTM layer with $100$ units, which generates the output trajectory (change in $x$ and $y$ direction, pressure, and stroke for every time step). During the one-shot inference mechanism, only the weights into the feedforward layer were adapted (blue).}
\label{Fig1}
\end{figure}

The model architecture is shown in Figure~\ref{Fig1}.
The input is the one-hot encoded vector of length $26$ for every time step, which is first processed in a fully-connected feedforward layer of $100$ neurons, followed by one LSTM layer of $100$ units, resulting in the output of pressure, stroke, and the relative change in $x$ and $y$ direction for every time step, which constitute the trajectory over time.
As we will see below, the fully-connected feedforward layer is highly useful when intending to recombine attractor dynamics in the RNN layer to quickly learn to generate untrained characters.
In preliminary experiments, the size of the architecture has been proven to be a good trade-off between model complexity and the quality of the generated outputs.

The model's weight parameters were trained for $500k$ training steps. 
We applied the mean squared error (MSE) loss function for training the model.
For gradient computation, we used Backpropagation Through Time \cite{werbos1990}. 
The weights of the network were optimized with Adam \cite{kingma_adam_2015} using default parameters (learning rate of 0.001, $\beta_1 = 0.9$ and $\beta_2 = 0.999)$.

After training, when presented with an untrained character trajectory and an unknown one-hot encoded label, the model generated a character trajectory that did, of course, not match the target trajectory. 
To probe the ability to freely generate new characters with only one example, we implemented the following \emph{one-shot inference} mechanism.
The gradient signals of the loss function were again propagated backwards through time. 
However, only the weights into the feedforward layer were adapted. 
The weights of the LSTM layer were not adapted at all.
This iterative process was repeated 13,000 times with a learning rate of 0.001.
As a result, the feedforward layer activities can be tuned such that the constant input activity flowing into the LSTM structure systematically activates those dynamic attractors and attractor successions that are best-suited to generate the novel character.  
From a cognitive perspective, this iterative inference process may be viewed as an imagination phase, in which the network essentially infers how to redraw the presented trajectory.

\section{Experiments}
In our experiments, we address the aforementioned four points of the character challenge. 
After learning, we probe if the network can re-generate a character, when presented with one example of a novel character, whether it can reliably identify such a novel character correctly as one particular novel type of character, 
and whether it can generate new variations of that novel character. 
Moreover, we probe if the network can generate completely new but related characters after being confronted with single examples of some new characters and we further analyse the hidden LSTM states.
Remember, that we applied varying human handwritten trajectories that are not always easily readable. Hence, it might be difficult to recognize some of the (realistically) generated letters by the model, too.

\subsection{Free Generation of Characters (after a single example)}
\begin{figure}[t!]
	\centering

	\includegraphics[width=.23\textwidth]{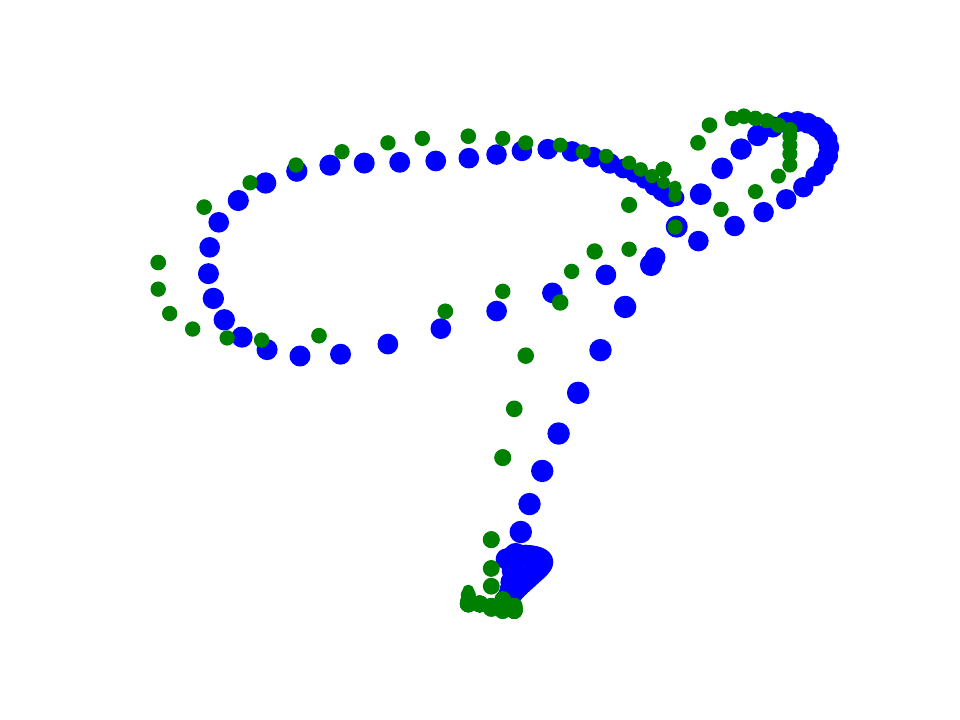} 				\includegraphics[width=.23\textwidth]{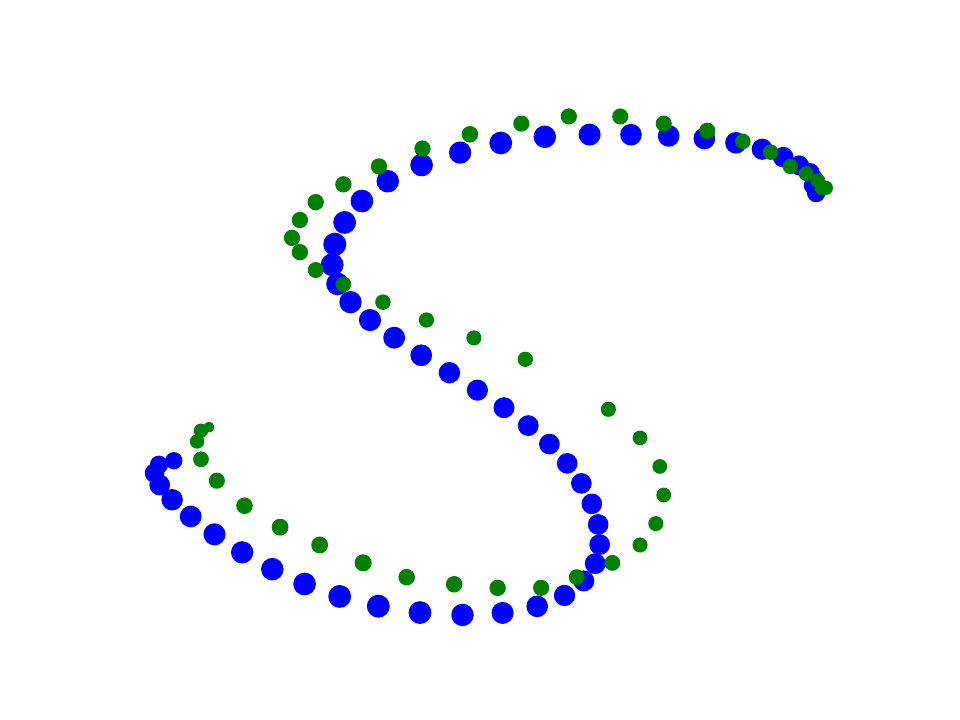}
	\includegraphics[width=.23\textwidth]{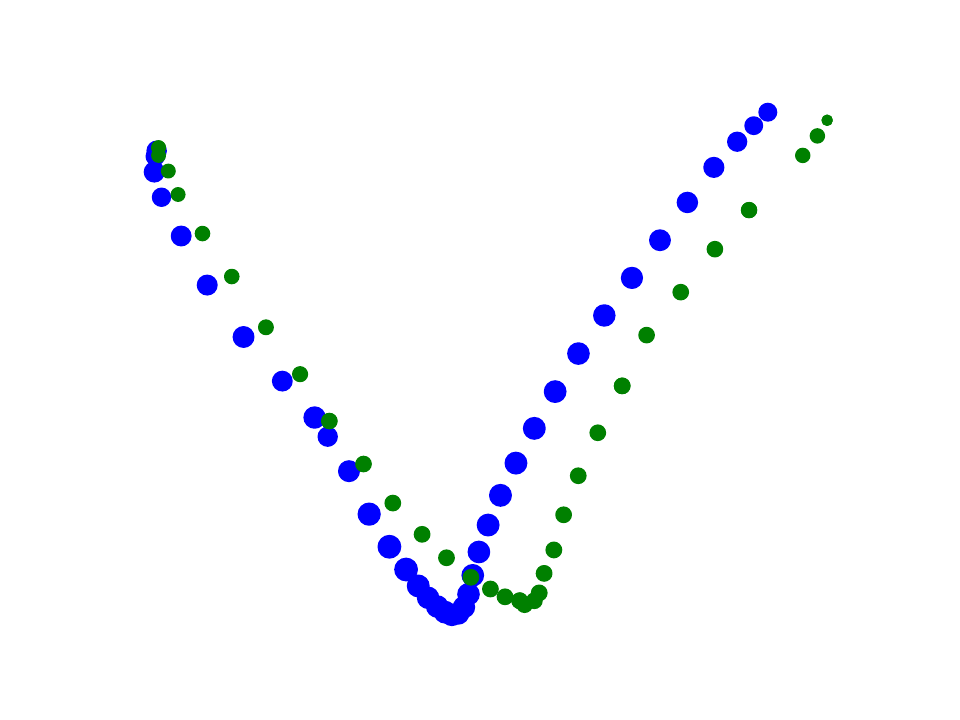}
	\\
	\includegraphics[width=.23\textwidth]{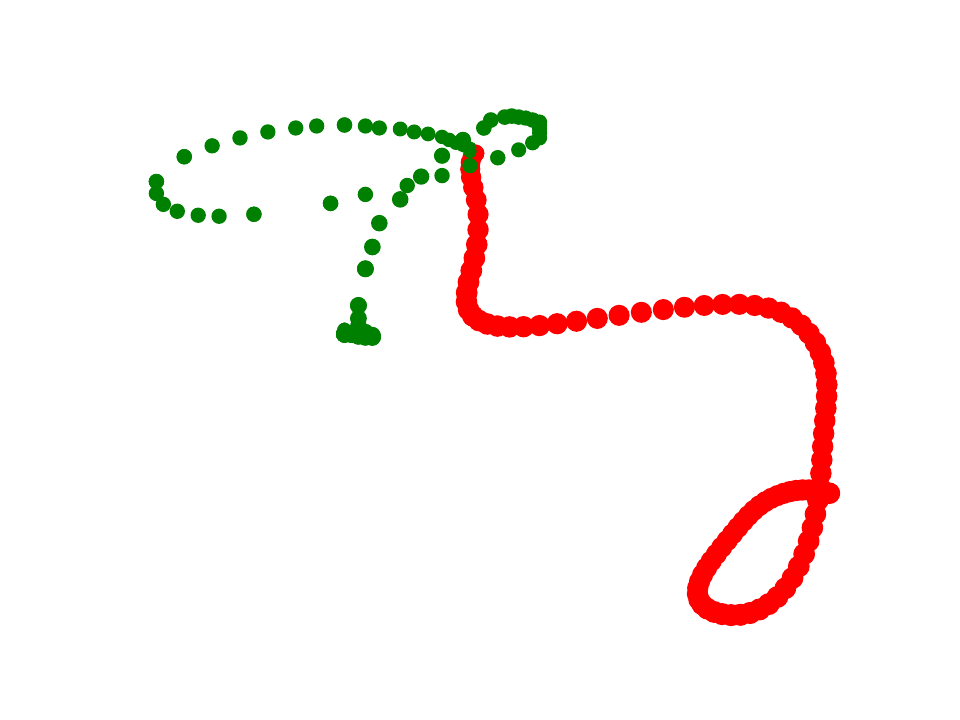} 				\includegraphics[width=.23\textwidth]{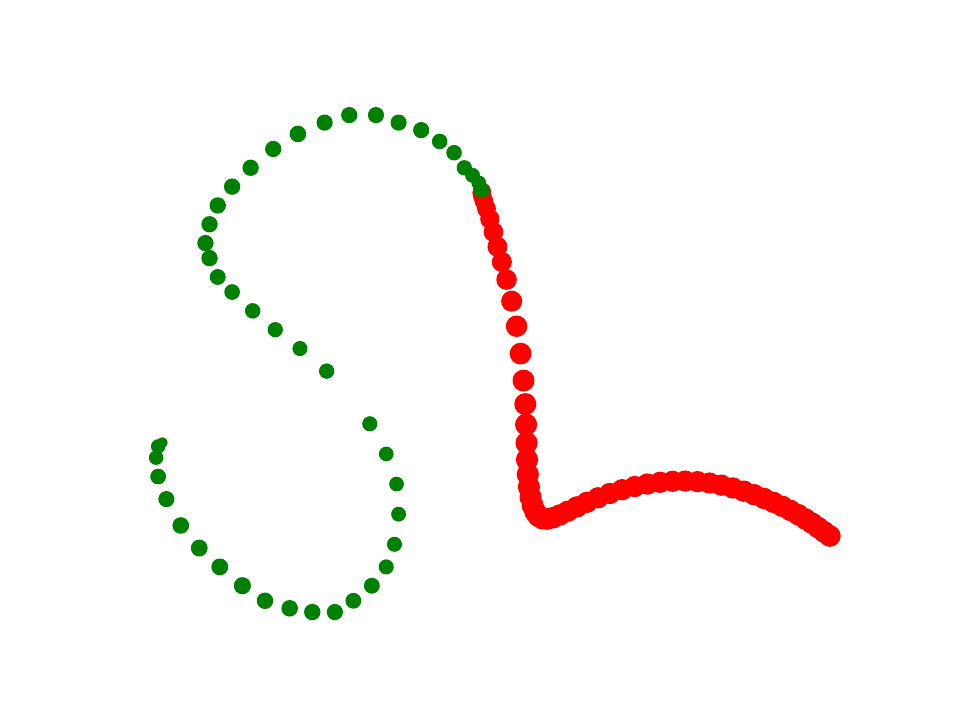}
	\includegraphics[width=.23\textwidth]{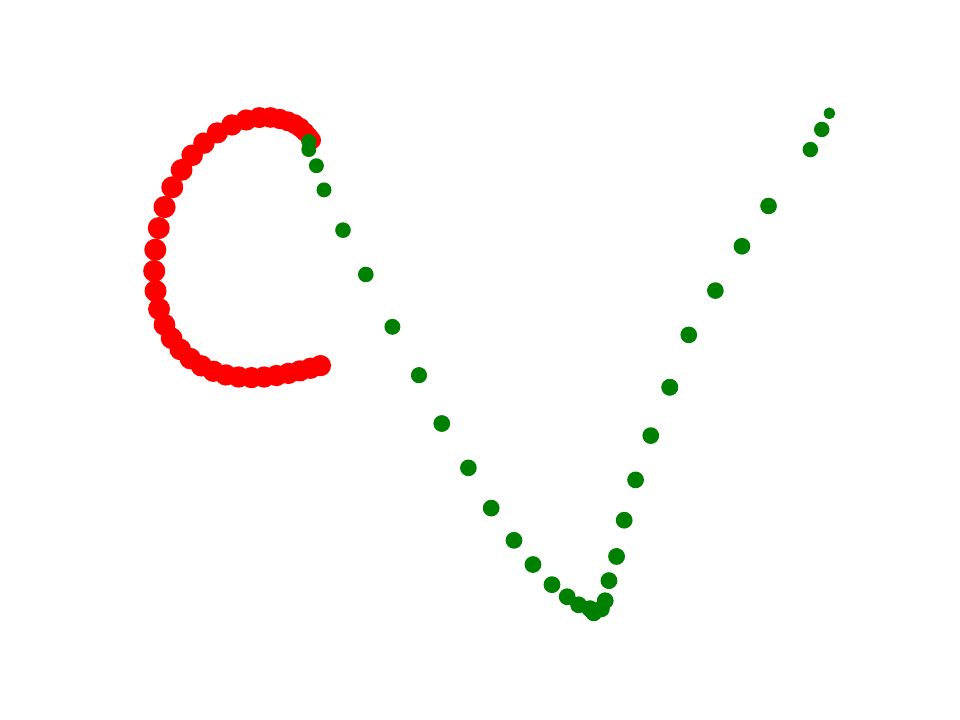}
	\\
	\includegraphics[width=.23\textwidth]{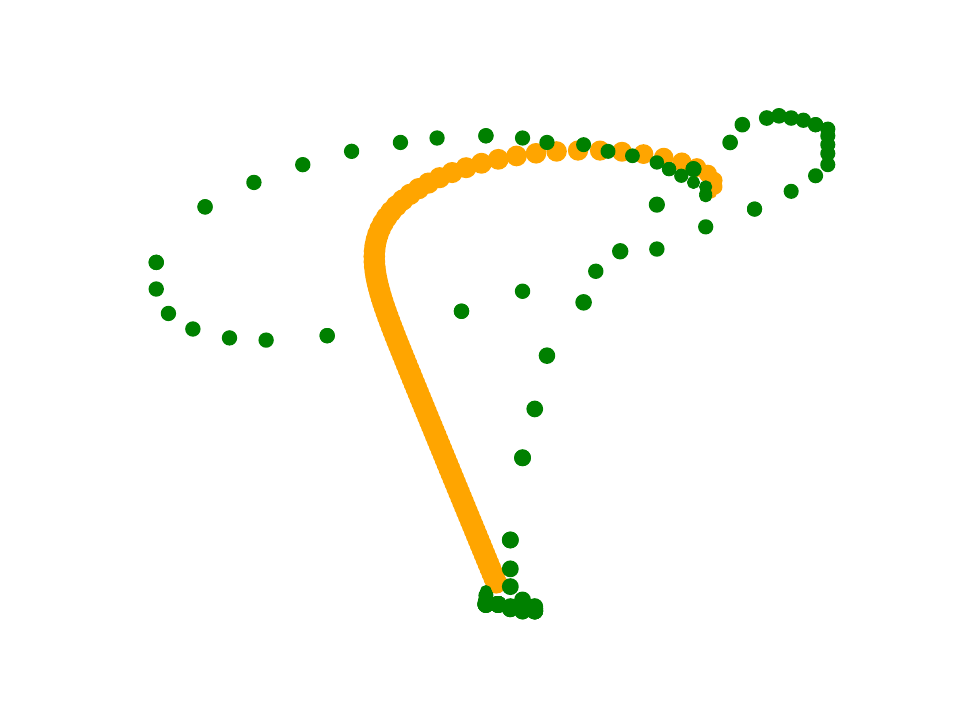} 				\includegraphics[width=.23\textwidth]{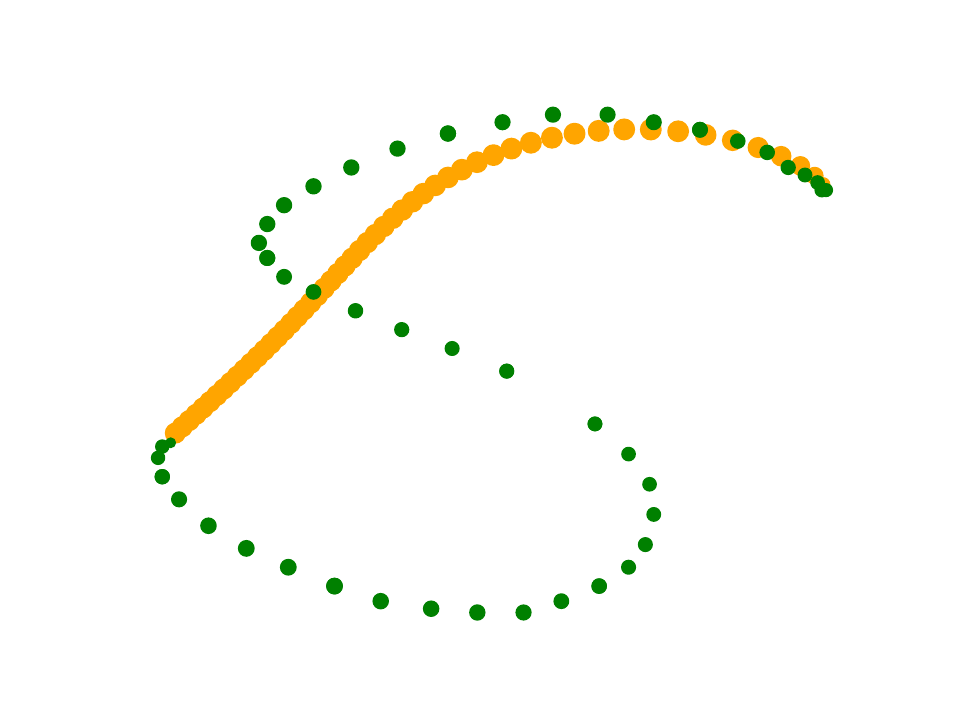}
	\includegraphics[width=.23\textwidth]{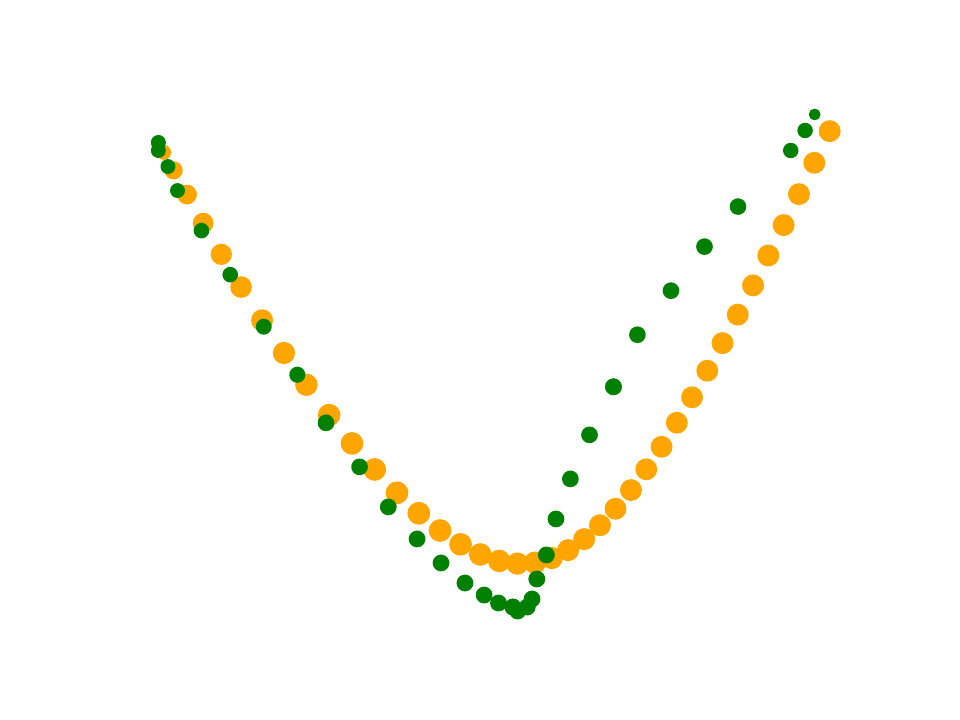}
  	\caption{Top: Three examples of re-generated trajectories after applying our one-shot inference mechanism (blue) given single examples of untrained characters (green), when applied in our character generation network architecture that was previously trained on other characters. 
  	Middle: Generated trajectories from unknown one-hot input vectors of the trained model without the one-shot inference mechanism (red). 
  	Bottom: Generated trajectories after the one-shot inference mechanism was applied to an untrained model (orange).}
 	\label{Fig2}
\end{figure}

Training against multiple training samples per character leads to a character model that generates one variant per character, essentially producing the mean of the encoded character concept. When presented with untrained inputs, the model is obviously not able to generate the correct trajectories (cf. Figure~\ref{Fig2}, red trajectories). But by means of our one-shot inference mechanism, our RNN architecture is indeed able to re-generate different untrained character trajectories 
(cf. Figure~\ref{Fig2}, blue trajectories).
Note that we do not provide or explicitly train our RNN architecture to encode basic motion primitives, that is sub-trajectories, as was done in \citeA{lake2015human}. 
Instead, our architecture has developed such sub-trajectories implicitly in its LSTM layer. 
As a result, it is able to compose the trajectory components it needs to generate the target trajectory by selective activation via the inferred, constant feed-forward layer activities, providing first hints that our architecture develops compositional, generative structures.

In contrast, an untrained model of the same architecture is not able to re-generate the characters via our one-shot inference mechanism, as shown in orange in Figure~\ref{Fig2}. 
The only exception for which the re-generated character looked similar to the original one in the untrained case was the `v' -- probably because of its simplicity. 
Even for this `v', the shape is less edged than in the original version. These results confirm that the training of other characters is indeed crucial, presumably because it fosters the development of character sub-trajectories that can be flexibly adapted and recombined to generate other characters of the alphabet. 

The importance of backpropagating the gradient onto the weights into the feedforward layer is further substantiated by the fact that similar attempts without the feedforward layer, like backpropagating the gradient onto constant one-hot mixture input vectors, have not been successful.

\subsection{Generation of New Samples of a Concept (after single example)}
\begin{figure}[t!]
\includegraphics[width=\textwidth]{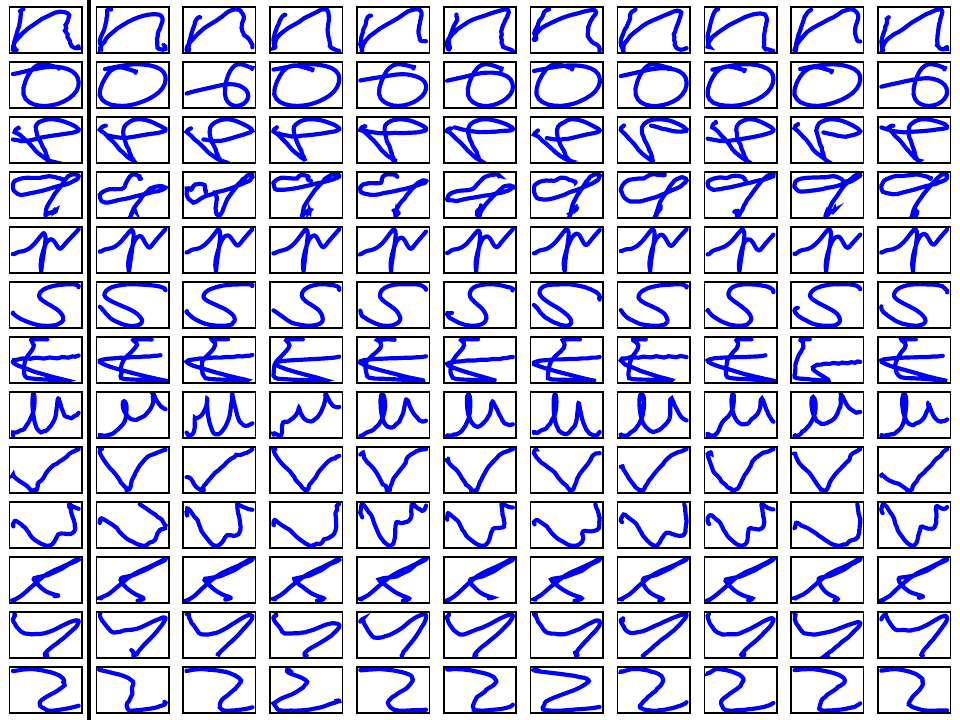}
\caption{The first column displays the re-generated character via one-shot inference, followed by generated variants via the addition of normally distributed noise to the inputs. 
} 
\label{Fig3}
\end{figure}

Next, we evaluate if the network architecture is able to generate new samples of a character that was learned from one single example presented to the model via our one-shot inference mechanism. 
After the adaptation of the weight vector into the feed-forward layer, we then added normally distributed noise ($M = 0$, $0.009 \leq SD \leq 0.07$) to the input label with 26 dimensions at every time step, allowing the network to create new instances of the presented target character. 
The generated variants shown in Figure \ref{Fig3} confirm that the network is indeed able to generate similar character variants.

\subsection{Identifying Novel Instances of a Concept (after single example)}
\begin{figure}[t!]
\begin{multicols}{4}
\centering
\includegraphics[width=.35\textwidth]{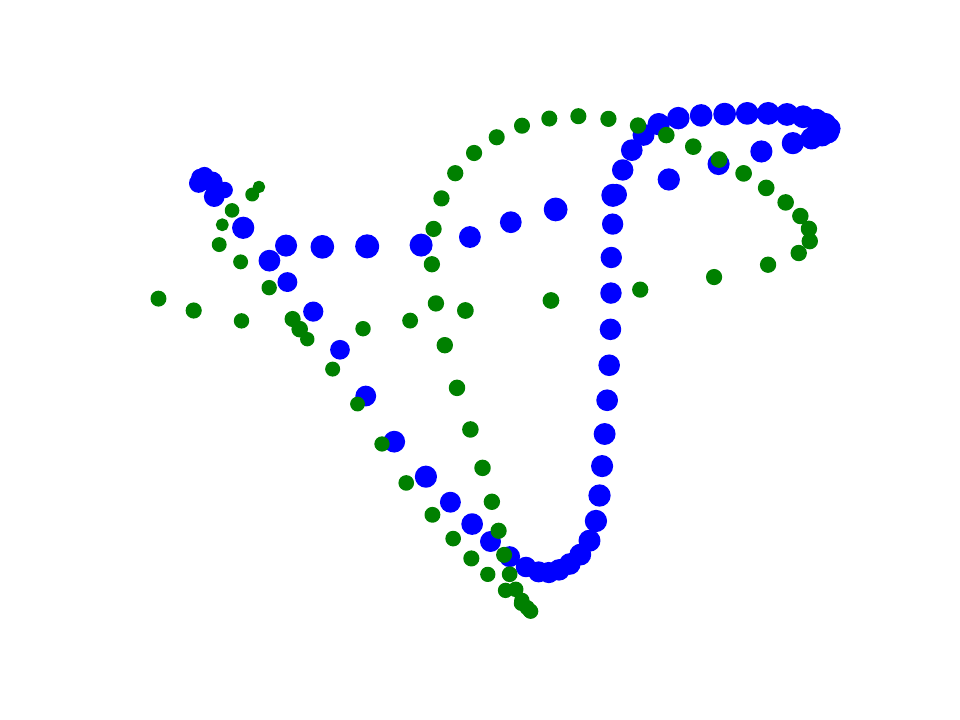}\\
\includegraphics[width=.08\textwidth]{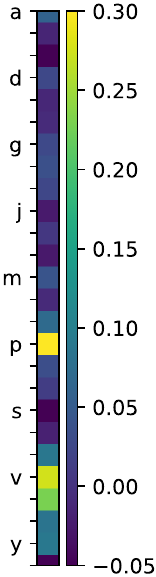}\\
\includegraphics[width=.35\textwidth]{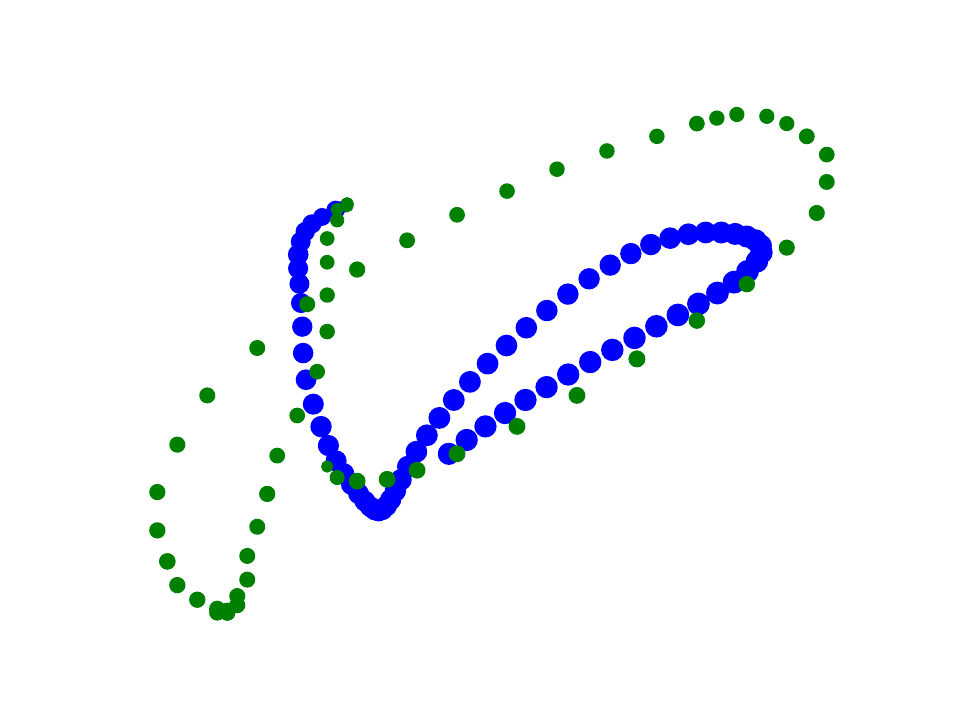}\\
\includegraphics[width=.08\textwidth]{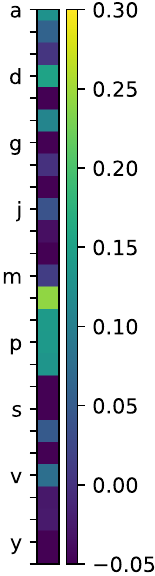}\\

\end{multicols}
\caption{Left: Variant of a `p' (green) and its reproduced trajectory through character type inference (blue) with the inferred generative input code. In 12 out of the 13 cases, the correct character type has been inferred. Right: Another type of `p' with different orientation (first stroke from top right to bottom left instead of top left to bottom right), resulting in an incorrect classification. Nevertheless, similarities between characters have been recognized by the model.} 
\label{Fig4}
\end{figure}

The next task of the character challenge is to distinguish a novel instance of an untrained character from other characters. 
We thus present the trained model first with one instance each for each untrained character (i.e. characters n - z), applying our one-shot inference mechanism with a distinct one-hot encoded input for each untrained character.
We then probe character type inference when confronted with a similar instance of one of those novel characters. 
During character type inference we do not provide any label (one-hot encoded) information, that is, we start with a zero vector of length $26$ as the input vector. 
We then backpropagate the gradient from the L2 loss onto that input vector, enforcing constant values for every time step. 
This iterative inference process is repeated $1k$ times with a learning rate of $0.01$ for every character.
As a consequence of this setup, the model is allowed to recombine the information of previously learned character codes, inferring a mixed label.  
The highest value of the inferred label determines the classification.
If the highest value is at the true position of the character, the classification is considered successful.
An example of a correctly and an incorrectly inferred input and the corresponding re-generated trajectory can be found in Figure~\ref{Fig4}. 
When applying very similar variants of the characters generated as explained in the last section, the model successfully infers the correct class in 12 out of the 13 cases. When using dissimilar variants of characters (for example print and script versions), the model is not able to determine the class reliably, which is not surprising because it has been shown only either one or the other variant. Nevertheless, the inferred inputs show that the system can detect similarities, since the correctly inferred `p' on the left in Figure~\ref{Fig4} shares some components with a `v', leading to an input vector that has high values both at the `p' and the `v' position. Even for the incorrectly classified `p' on the right, the high values at the `n', `o', `p', `q', and `d' positions seem reasonable given their shared components.

\subsection{Generation of Entirely New Concepts (after single examples of some concepts)}
Finally, we investigate whether the system is able to generate novel characters in a somewhat innovative manner, ideally generating characters that do not exist but that nonetheless look like plausible characters. 
We realize this aspect by investigating the effects of blending two characters. 
Again, we use the trained model with the feedforward layer input weights for the characters n - z optimized via one-shot inference. 
We present the resulting model with blending input vectors with two non-zero values that sum up to one. 
In a sense, this input vector instructs the model to generate a trajectory that expresses a compromise between two character trajectories, mixing and blending sub-trajectories of each character. 
Figure~\ref{Fig5} shows that our network trajectory indeed generates innovative character blendings. 
The observable smooth blending transitions from one character to the other underline the compositional recurrent codes that developed in the LSTM layer. A video that illustrates the blending between different characters can be found here: \url{https://youtu.be/VyqdUxrCRXY}

\begin{figure}[t]
\includegraphics[width=\textwidth]{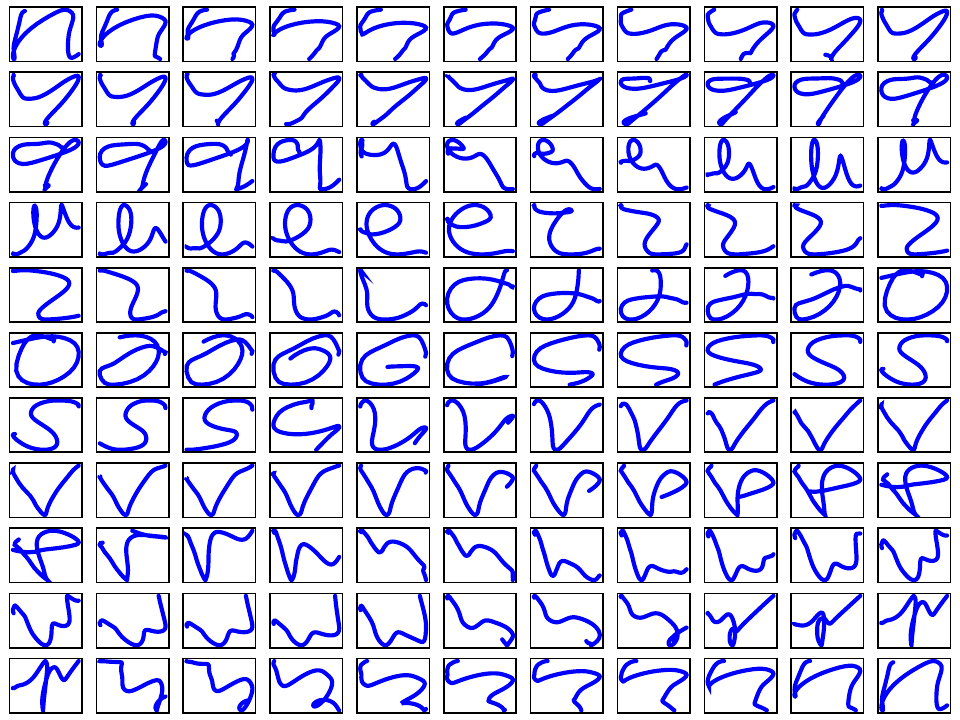}
\caption{Blending of single examples of untrained characters in ten steps. The second column shows blendings of 90 \% of the first and 10 \% of the second character, the third column 80 \% of first and 20 \% of second character and so on. Note that blending for trained characters (a - m) looks conceptually similar. 
} 
\label{Fig5}
\end{figure}

\subsection{Analysis of Hidden LSTM States}
\begin{figure}[t!]
\centering
\includegraphics[width=.48 \textwidth]{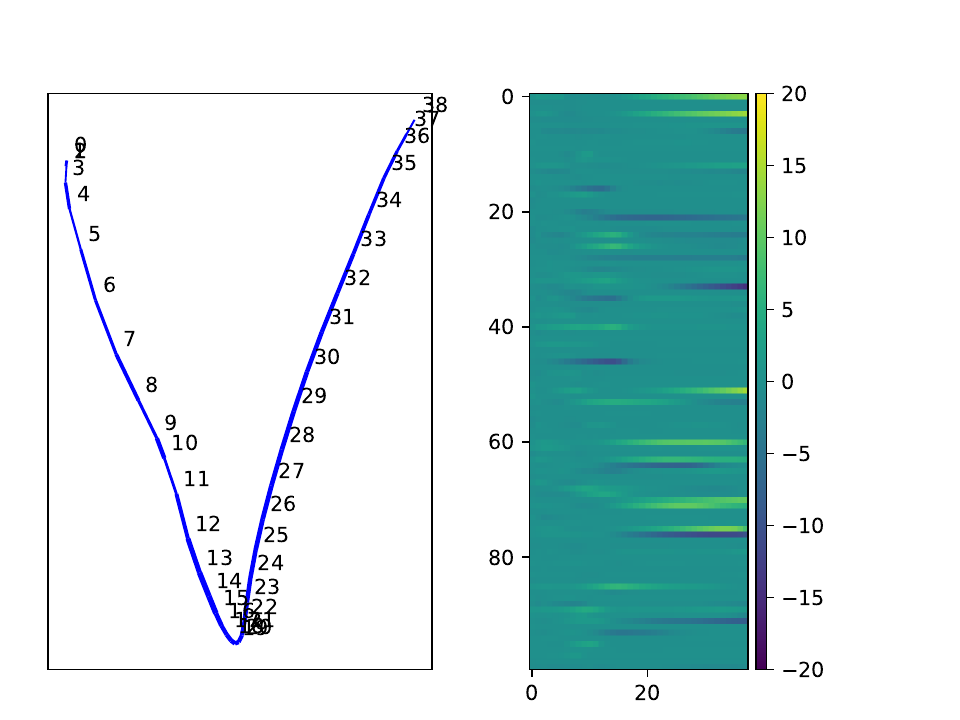}
\includegraphics[width=.48 \textwidth]{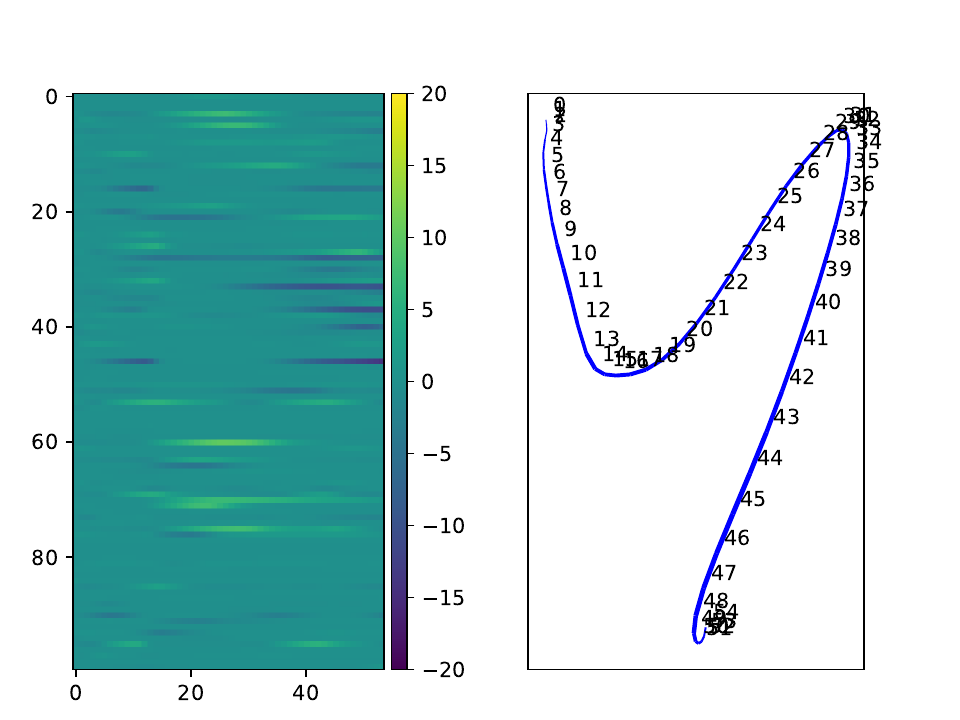}
\caption{Re-generated trajectories with time steps and the values (represented by color) of the corresponding hidden cell states c of the LSTM cells over time.} 
\label{Fig6}
\end{figure}

To shed more light on the nature of the encodings that have developed in the hidden LSTM cell states, we further analyzed the neural activities while generating particular character trajectories.
Hidden neural cell state activities $c$ of the LSTM layer and the corresponding trajectories are plotted in Figure~\ref{Fig6}.
Although only exemplarily, the analysis confirms that similar sub-dynamics unfold when similar sub-trajectories are generated: 
For the character `v', the downward (approx. steps 1-16) and upward (approx. steps 21-37) strokes reveal distinct but partially stable patterns in the LSTM cell states.
Most interestingly, a closely related pattern can be detected for the first part of the trajectory of the character `y' (approx. steps 2-30), essentially drawing a similar `v' shaped sub-trajectory. 

\begin{figure}[t!]
\centering
\includegraphics[width=.47\textwidth]{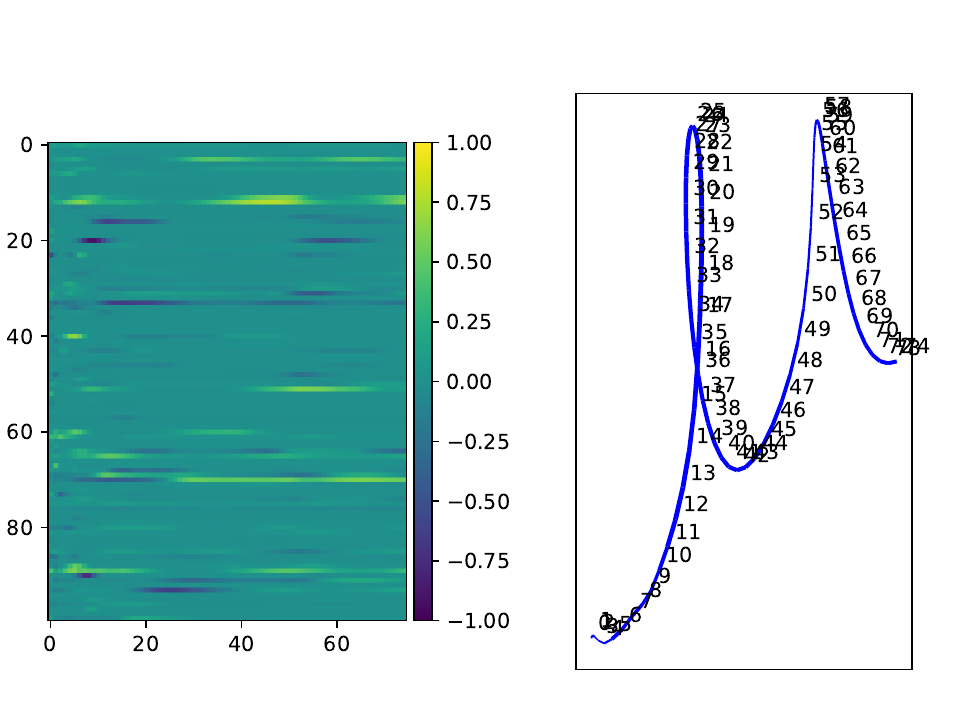}
\hfill
\includegraphics[width=.47\textwidth]{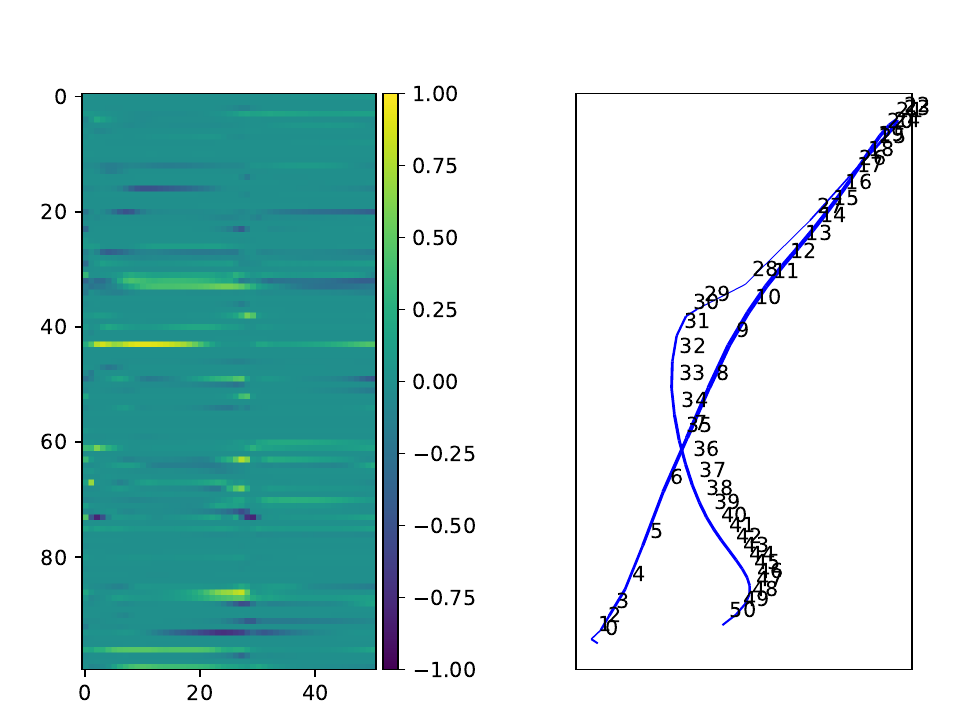}
\caption{Re-generated trajectories with time steps and the values (represented by color) of the corresponding hidden states h of the LSTM cells over time.} 
\label{Fig7}
\end{figure}

Figure~\ref{Fig7} shows some exemplary hidden states $h$ of the LSTM layer. 
When generating the character `u' a pattern repetition can be detected for the two upwards-downwards motions. 
For the character `x', distinct diagonal upwards, jump, and cross-diagonal downwards patters are visible in the hidden states.

\section{Conclusion}
In a review on the models having attempted to solve the character challenge until 2019, Lake and colleagues \citeA{lake2019omniglot} stated that except for the one-shot classification task, there has not been a lot of progress on the other tasks. They expressed their hopes that `researchers will take up the challenge of incorporating compositionality and causality into more neurally-grounded architectures’. 
The current paper provides important insights regarding efficient learning, the emergence of compositional encodings and recombinations thereof, and the integration of a type of imagination and planning into RNNs.

Our generative feed-forward-LSTM model, combined with a one-shot inference mechanism, was able to meet the character challenge. Deep learning methods are usually bottom-up methods that need a large number of training examples. Lake and colleagues \citeA{lake2015human} applied a top-down approach by giving their program information about the existence of components, like strokes, half-circles and so on.
Our approach was able to re-generate unseen character trajectories over time from just one example of a novel character, without providing any a priori structured motor primitives. 
This indicates that the system combined the knowledge of previously learned characters in an innovative manner to generate untrained characters, providing evidence that LSTM networks can indeed (i) partition time-series data implicitly into components, which encode sub-trajectories, and (ii) recombine them in a compositional manner to efficiently learn new characters.

The network and inference mechanisms were furthermore able to classify different variants of a character as belonging to the same one, as long as the presented trajectory variants were closely related. 
However, when the network was presented, for example, with a print `t', it was not able to classify the trajectory of a script `t' -- which starts at the bottom and continues upward instead of starting at the top, continuing downward.
This makes sense conceptually because our model encodes the motor trajectory in a recurrent, generative manner. 
It does not encode the actual image of the character that was generated. 
As a consequence, it classifies trajectory similarities, not image similarities. 
This corresponds to the fact that humans may classify both a script and a print `t' as the character `t' but indeed need to invoke very different motor programs when generating the one or the other, and switching between both styles comes with effort. 
Accordingly, one-shot classification is only possible for similar trajectory variants with the presented method. 
In the future, we intend to enhance our model with an encoder-decoder-oriented convolutional module, which may indeed interact with our trajectory generation module and the one-hot encoded classification layer, which we used as input to our generative architecture. 
 
A further interesting result is that by using the learned components from the known characters, the model generated new examples of a particular type of character and even novel but plausibly looking character trajectories by blending previously seen ones in a somewhat innovative, smooth manner.  
Additionally, the visualization of recurrent hidden states showed similar patterns for characters that share similar sub-trajectories, providing interesting insights regarding the explainability of LSTMs, indicating the emergence of compositional dynamic attractor patterns within LSTM's hidden states. 
Further analyses should be conducted to shed additional light on the nature of these dynamic patterns.  

Overall, these results provide strong evidence that LSTM networks tend to develop kinds of compositional encodings, which may be reused to generate untrained, but related trajectories in fast and innovative manners. 
Such combinatorial generalization abilities are of course not restricted to letter trajectories, but can be applied to all time series patterns. They are of major significance, since they seem to be a key ingredient of human intelligence, which is why AI researchers have been interested in combinatorial abilities since the origins of AI \cite{battaglia2018relational}.
The awareness and utilization of these compositional abilities of RNNs will hopefully inspire future research and may be an essential aspect towards bridging the gap between human and machine intelligence.

\subsection*{Acknowledgements}
\begin{footnotesize}
The results of this work were produced with the help of the GPU cluster of the BMBF funded project Training Center for Machine Learning (TCML) at the Eberhard Karls Universit\"at T\"ubingen, administered by the Cognitive Systems group. We especially thank Maximus Mutschler who is responsible for the maintenance of the cluster.
\end{footnotesize}

\bibliographystyle{splncs04}
\bibliography{references}
\end{document}